\documentclass[pdftex]{nolta}

\Vol{2}%
\No{3}%
\Year{2024}%
\Month{4}%
\title{Proposed modified computational model for the amoeba-inspired combinatorial optimization machine}
\AUTHOR*{%
\author{Yusuke Miyajima}{1}\orcid{0009-0001-4895-8643}, 
\author{Masahito Mochizuki}{1}\orcid{0000-0002-0128-0580}
}
\AFFILIATE{%
\affiliate{Department of Applied Physics, Waseda University, Okubo, Shinjuku-ku, Tokyo 169-8555}{1}
}
\received{10}{27}{20XX}
\revised{12}{29}{20XX}
\published{7}{1}{20XX}
\begin{document}
\begin{abstract} 
A single-celled amoeba can solve the traveling salesman problem through its shape-changing dynamics. In this paper, we examine roles of several elements in a previously proposed computational model of the solution-search process of amoeba and three modifications towards enhancing the solution-search preformance. We find that appropriate modifications can indeed significantly improve the quality of solutions. It is also found that a condition associated with the volume conservation can also be modified in contrast to the naive belief that it is indispensable for the solution-search ability of amoeba. A proposed modified model shows much better performance.
\end{abstract}

\maketitle 
\section{Introduction}
In the modern information society, conventional Neumann-type computers face a serious problem of increased complexity and cost in computation. While the information processing capacity of computers is predicted to reach its limits because of the limit of Moore's law and the von Neumann bottleneck, the required computing power is increasing exponentially. Under these circumstances, new efficient domain-specific computational architectures beyond the von Neumann type are demanded \cite{Hennessy17}. The serious problem particularly appears in combinatorial optimization problems \cite{Garey79}. Recently, the Ising machine has been proposed as a solution and its practical applications are being investigated \cite{Aramon19, Johnson11, Goto19, Inagaki16, Yamaoka16}. There, the combinatorial optimization problem is mapped onto the ground-state search of the Ising model, which is originally a mathematical model of magnetic materials \cite{Kadowaki98, Lucas14}. On the other hand, research has been intensively conducted to construct combinatorial optimization machines that mimic the information processing of living organisms. This is because organisms often perform sophisticated information processing such as image recognitions, sound recognitions, and optimizations with low energy consumption.

One of the simplest and most useful machines of such kinds is a computing technology inspired by the information processing of the amoeboid organism \cite{Adamatzky16, Aono20}. This technology exploits the dynamics of unicellular slime mold, which transforms its body in response to the surroundings for survival. Specifically, it exploits the behaviors of the body to stretch for getting food and to shrink for avoiding light \cite{Kessler82}. As demonstrated in the previous experimental study, unicellular slime molds are able to solve optimization problems such as mazes \cite{Nakagaki00, Tero10}, satisfiability problems \cite{Aono07}, and traveling salesman problems (TSP) \cite{Aono09, Zhu13, Iwayama16}. The superior solution-search capability derives from the shape-changing dynamics of amoeba, where the law of conservation of total volume is satisfied \cite{Iwayama16}. In other words, slime molds have high ability of the information processing despite being unicellular organisms and lacking a central information-processing mechanism. This situation has attracted attention as a model for an autonomous decentralized parallel computing architecture of high efficiency beyond the von Neumann architecture. However, motion of slime molds is as slow as $10^{-5}$ m/s, making it difficult to use them as practically usable machines. Therefore, research interest has focused on attempts to implement the information-processing processes of slime molds with physical devices as amoeba-inspired combinatorial optimization machines \cite{Kasai13, Aono13, Aono15, Takeuchi19, Hara-Azumi20, Saito20a}. For simplicity, we will refer to slime mold as amoeba hereafter.

To realize the amoeba-inspired machines, a computational model (algorithm) for the solution-search process of amoeba should be constructed first \cite{Aono12, Zhu18, Saito20b}. An important model has been proposed by Aono and his coworkers, which is named Amoeba TSP algorithm \cite{Zhu18}. Its physical implementation using analogue electronic circuits has been demonstrated by a subsequent experiment \cite{Saito20a}. The model has turned out to describe well the amoeba dynamics and resulting solution-search ability. However, we do not have sufficient knowledge about roles of elements constituting the model on its solution-search ability. For example, the fluctuations expected to exist in the amoeba dynamics are taken into account using uniformly distributed random numbers within a certain range. However, the necessity of the fluctuations for the optimization ability is not trivial. Moreover, it has not been clarified how the optimization performance is affected if we change the distribution of random numbers. Insights into the role of each element in the model are expected to provide the guide for constructing the amoeba-inspired machines with higher solution-search ability.

In this paper, we study roles, effects, and (un)necessities of several elements of the Amoeba TSP algorithm by examining how modifications of the elements affect the optimization performance of the algorithm. It is found that appropriate modifications can significantly improve the optimization performance. On the basis of these examinations, we propose a modified model named Improved Amoeba TSP algorithm, which turns out to provide much better solutions in terms of the number of iterations and the route length. It has also turned out that even a condition originating from the volume conservation law assumed in the original model can be modified for the high solution-search ability in contrast to the naive belief, which may release us from the related constraints against the physical implementation.

This paper is structured as follows. In Section 2, a formalism of the TSP proposed by Hopfield and Tank is introduced \cite{Hopfield86}. Section 3 describes the concept of amoeba-inspired combinatorial optimization machine for solving the TSP. Specifically, an experiment with a biological slime mold and a computational model named Amoeba TSP algorithm are introduced. In Section 4, the parameters and conditions of the experiment and numerical simulations are described. In Section 5, we argue three elements of the model that characterize the Amoeba TSP algorithm and examine how modifications of these elements affect the optimization performance. In Section 6, a computational model named Improved Amoeba TSP algorithm is proposed. Section 7 is devoted to the summary.

\section{Formalism of the traveling salesman problem}
We first present a formalism of the TSP \cite{Applegate06}. The TSP is a problem to find the shortest route for a given map which satisfies the condition that every city must be visited (only) once. When the number of cities is $n$, the number of candidate solutions is $(n-1)! /2$, which increases exponentially with increasing $n$. The TSP is known as one of the NP-hard problems, that is, no algorithm can solve this problem within polynomial computational time. Therefore, it is a popular approach in practical cases to search an approximate solution by heuristic algorithms. In our study, we attempt to obtain an approximate solution using the amoeba-inspired combinatorial optimization algorithm.

We start with a brief introduction of the formalism proposed by Hopfield and Tank. For the $n$-city TSP, $n^2$ binary variables $x_{Vk}(=0, 1)$ are used to represent the routes, where the indices $V(=1, 2, \cdots, n)$ and $k(=1, 2, \cdots, n)$ represent a labeling number of the cities and a visiting order of the cities, respectively. Here, $x_{Vk}=1 (0)$ means that the city labeled as $V$ is (is not) visited at the $k$th order. The cost function to be optimized in the searching procedure is defined as, 
\begin{equation}
\label{eq:CostFunc}
E(\{x_{Vk}\}) = -\frac{1}{2} \sum_{V=1}^n \sum_{U=1}^n \sum_{k=1}^n \sum_{l=1}^n W_{Vk, Ul}
\;x_{Vk} \;x_{Ul}. 
\end{equation}
Two constraints of the TSP are given by the following equations,
\begin{equation}
\label{eq:Constraints}
\sum_{k=1}^{n} x_{Vk}=1, \hspace{1cm} \sum_{V=1}^{n} x_{Vk}=1.
\end{equation}
The former constraint prohibits revisiting once-visited cities, and the second constraint prohibits visiting more than one city simultaneously. The TSP is reduced to a problem to search a combination of $n^2$ binary variables $\{x_{Vk}\}$ which minimizes the cost function in Eq.~(\ref{eq:CostFunc}) with satisfying the constraints. Here, the cost $W_{Vk, Ul}$ is concerned with the distance between two cities and the penalty for violation of the constraints, which is defined by,
\begin{equation}
\label{eq:Costs}
W_{Vk, Ul} = \left\{
\begin{array}{ll}
-\lambda & (V=U \land k \neq l), \\
-\mu & (V \neq U \land k=l), \\
-\nu \cdot d(V, U) & (V \neq U \land | k - l |=1),\\
-\nu \cdot d(V, U) & (V \neq U \land k=n \land l=1),\\
-\nu \cdot d(V, U) & (V \neq U \land k=1 \land l=n),\\
0 & (\rm{otherwise}).
\end{array} \right.
\end{equation}
Here $\lambda$, $\mu$ and $\nu$ are positive parameters, and $d(V, U)$ represents the distance between two cities $V$ and $U$. The parameters $\lambda$ and $\mu$, respectively, give costs for the penalty when the constraints in Eq.~(\ref{eq:Constraints}) are violated, while $\nu$ gives a cost proportional to the distance between two cities visited successively. The fourth and the fifth costs are concerned with the distance between the final and initial cities (The salesman must come back to the initial city finally). The value of the cost function is proportional to the route length when the constraints are satisfied. Note that the cost function in this formulation does not explicitly include the penalty terms for satisfying the constraints such as $\sum_{k=1}^{n} x_{Vk} - 1$ and $\sum_{V=1}^{n} x_{Vk} - 1$ in contrast to other optimization machines, e.g., Ising machines.

In this study, we evaluate the results obtained by the amoeba-based optimization machine and the computational models (algorithms) in terms of the following three criteria,
\begin{itemize}
\item Success rate for finding an approximate solution within the given time or number of iterations,
\item Average of time or number of iterations required to find an approximate solution,
\item Average of the normalized route length for the obtained solutions
\end{itemize}
The averages in the latter two are taken over all the successful trials of the solution search. The normalized route length of an obtained solution is defined as $R_{\rm calc}/R_{\rm est}$, where $R_{\rm calc}$ and $R_{\rm est}$ are the route length of the obtained solution and the estimated mean route length, respectively. A smaller average of $R_{\rm calc}/R_{\rm est}$ means that the amoeba-based optimization machine or the computational model is capable of finding a solution with a shorter route length.

\section{Amoeba-inspired combinatorial optimization machine}
\subsection{Experiment}
\begin{figure*}[thb]
\includegraphics[scale=0.8]{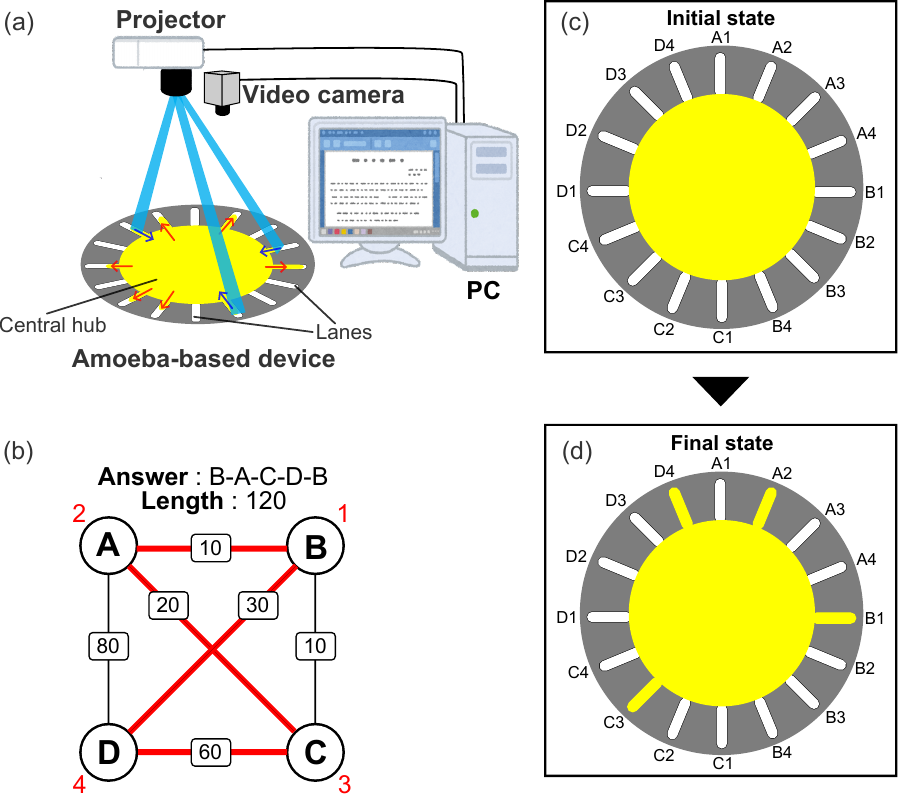}
\caption{(a) Experimental setup of the amoeba-based combinatorial optimization machine. (b) Example of a map of the 4-city TSP. In this experiment, the cities are labeled by alphabets, A, B, C and D. (c),~(d) Initial and final states of the amoeba-based device in the experiment. The final state shown in (d) represents a solution corresponding to a traveling route in order of B $\rightarrow$ C $\rightarrow$ D $\rightarrow$ A. The red lines and the red numbers in (b) represent the traveling route and the visiting order, respectively.}
\label{fig:experiment}
\end{figure*}

Aono and coworkers experimentally demonstrated that the shape-changing dynamics of amoeba induced by environmental stimuli, i.e., nutrient and light, can be used for efficient searches for approximate solutions of the TSP \cite{Aono09}. Figure~\ref{fig:experiment}(a) shows schematics of the amoeba-based device and the optical feedback system used in the experiment, while Figs.~\ref{fig:experiment}(c) and \ref{fig:experiment}(d) show initial and final states of the amoeba-based device. The device is set on an agarose plate with nutrient. The optical feedback system consists of a PC, a projector and a video camera.
The device is composed of a central hub part and $n^2$ lanes to represent solutions of the TSP. Each lane corresponds to one continuous variable $X_{Vk} \in [0, 1]$ for $V=1, 2, \cdots , n$ and $k=1, 2, \cdots , n$ and thus is referred to as the lane $Vk$. The body of amoeba in a lane is called branch. The variable $X_{Vk}$ is defined as the ratio of area occupied by the amoeba to the total area in the lane $Vk$, which is referred to as length of the branch in the lane $Vk$. The cost function with binary variables $\{x_{Vk}\}$ in Eq.~(\ref{eq:CostFunc}) is adopted in the experiment. The continuous variables $X_{Vk}$ are translated to the binary variable $x_{Vk}$ where $x_{Vk}$ equals to 1 (0) when $X_{Vk}$ equals to (is less than) unity. Namely, $x_{Vk}$ takes 1 only when the lane is fully occupied by the branch. We read out a solution $\{x_{Vk}\}$ of the TSP from a state of the branches of amoeba $\{X_{Vk}\}$ in this manner in the experiment. 

The optical feedback system illuminates the lanes and updates the state of the branches so as to minimize the cost function. The value of $X_{Vk}$ is measured by the video camera, and $L_{Vk}$ defined by Eq.~(\ref{eq:expL}) is calculated from all the values of $X_{Vk}$ by the PC. Here $L_{Vk}$ is a function of an integer variable $t$ representing the number of time step. A time step corresponds to six seconds.
\begin{equation}
\label{eq:expL}
L_{Vk}(t+1) = 1 - \sigma_{1000, -0.5} \left( \sum_{U=1}^n \sum_{l=1}^n W_{Vk, Ul}\;
\sigma_{35, 0.6}(X_{Ul}(t)) \right). 
\end{equation}
Here $\sigma_{\gamma, \theta}$ is a sigmoid function defined by, 
\begin{equation}
\label{eq:Sigmoid}
\sigma_{\gamma, \theta} (x) = \frac{1}{1+\exp(-\gamma(x-\theta))}.
\end{equation}
The lane $Vk$ will (will not) be illuminated by the projector when $L_{Vk}$ is greater (less) than 0.5. We call a combination of the step functions $\theta (L_{Vk}-0.5)$ for all the lanes as the illumination pattern for each step. The amoeba basically elongates its branches in non-illuminated lanes to efficiently absorb the nutrient from the agarose plate. On the contrary, the amoeba basically contracts its branches in illuminated lanes to avoid the light stimuli. The illumination is a penalty to prevent amoeba from choosing solutions with significantly long route lengths or those violating the constraints in Eq.~(\ref{eq:Constraints}).

In the initial state, the amoeba occupies only the hub part of device and does not occupy any lanes at all, that is, $X_{Vk}=0$ for all sets of $Vk$  [Fig.~\ref{fig:experiment}(c)]. The operation for calculating the illumination pattern $\{ \theta (L_{Vk}-0.5) \}$ and updating a state of the branches $\{X_{Vk}\}$ by illumination or non-illumination is iterated to minimize the cost function. Finally, only the $n$ lanes are fully occupied with amoeba's branches and the others are empty, that is, $X_{Vk}$ equals to 1 for $n$ sets of indices $Vk$ [Fig.~\ref{fig:experiment}(d)]. Eventually, an approximate solution is obtained [Fig.~\ref{fig:experiment}(b)].

According to the results of this experiment \cite{Zhu18}, the approximate solutions are obtained for almost all the trials and the average time required to obtain a solution increases linearly to the number of cities $n$. It has also turned out that the average of the normalized route lengths $R_{\rm calc}/R_{\rm est}$ takes 0.9 almost irrespective of $n$. Although the number of candidate solutions grows exponentially with increasing $n$, the amoeba is able to efficiently find an approximate solution with a reasonable route length. The detailed observation provides us an insight that this high solution-search ability of the amoeba originates from the fluctuating dynamics. The amoeba sometimes elongates (contracts) its branches even though the corresponding lane is (is not) illuminated. These slightly fluctuating responses enable to escape from local optimal solutions and to explore wider space of solutions in the searching process. However, it takes several ten minutes to find an approximate solution for real amoebas. This time scale is too long for practical use. Therefore, it is necessary to implement this high solution-searching capability of amoeba on a physical combinatorial optimization machine.

\subsection{Computational model: Amoeba TSP algorithm}
A computational model named Amoeba TSP algorithm has been proposed to reproduce the results of above experiment \cite{Zhu18}. This model describes the time evolution of amoeba's branches in the lanes $X_{Vk}(t)$ by the following equation,
\begin{equation}
\label{eq:modelX}
X_{Vk}(t+1)=\left\{
\begin{array}{ll}
X_{Vk}(t) - O_{Vk}(t+1) + \xi_{Vk}(t+1) & (L_{Vk}(t+1) > 0.5), \\
X_{Vk}(t) + I_{Vk}(t+1) + \xi_{Vk}(t+1) & (L_{Vk}(t+1) \leq 0.5).
\end{array} \right.
\end{equation}
In this mode, the value of $X_{Vk}$ can take an arbitrary real number and is not restricted to $[0,1]$ unlike the experiment. The state of the branches $\{ X_{Vk} \}$ is updated according to this equation. Here the variables $O_{Vk}$ ($I_{Vk}$) are amounts of contraction (elongation) of a branch in the lane $Vk$ when the lane is (is not) illuminated. The variables $\xi_{Vk} \in [-\delta, \delta]$ are uniform random numbers describing intrinsic fluctuations in the dynamics of amoeba branches.

First, $L_{Vk}(t+1)$ is calculated from $X_{Vk}(t)$ using Eq.~(\ref{eq:expL}) with the costs $W_{Vk, Ul}$ in Eq.~(\ref{eq:Costs}) for all the sets of $Vk$. Next, the branch lengths $X_{Vk}(t)$ are updated to $X_{Vk}(t+1)$ using Eq.~(\ref{eq:modelX}) based on the values of $L_{Vk}$. Iterating the calculations of the illumination pattern $\{\theta(L_{Vk}-0.5)\}$ and updating the state of the branches $\{X_{Vk}\}$ alternately, we obtain an approximate solution of the TSP. We perform these two procedure based on Eq.~(\ref{eq:expL}) and Eq.~(\ref{eq:modelX}) for all lanes as one iteration step. Here, the number of iterations is given by a integer $t$ and the variables updated in every iteration step are defined as functions of $t$.

The variables $O_{Vk}(t)$ and $I_{Vk}(t)$ are defined by,
\begin{equation}
\label{eq:modelO}
O_{Vk}(t+1)=\left\{
\begin{array}{ll}
2\Delta^{\rm out} \sigma_{20, 0.6}(X_{Vk}(t)) & (L_{Vk}(t+1) > 0.5) \\
0 & (L_{Vk}(t+1) \leq 0.5)
\end{array} \right.
\end{equation}
\begin{equation}
\label{eq:modelI}
I_{Vk}(t+1)=\left\{
\begin{array}{ll}
\left( \Delta^{\rm in} + \sum_{V=1}^n \sum_{k=1}^n O_{Vk}(t+1) + S(t) \right) / L^{\rm off}(t+1) & (L^{\rm off}(t+1) > 0) \\
0 & (L^{\rm off}(t+1) = 0) \\
\end{array} \right.
\end{equation}
with
\begin{equation}
\label{eq:modelS}
S(t+1)=\left\{
\begin{array}{ll}
0 & (L^{\rm off}(t+1) > 0) \\
S(t) + \Delta^{\rm in} + \sum_{V=1}^n \sum_{k=1}^n O_{Vk}(t+1) & (L^{\rm off}(t+1) = 0) \\
\end{array} \right.
\end{equation}
Here $\Delta^{\rm out}$ and $\Delta^{\rm in}$ are parameters representing the degree of the contraction and the elongation of branches. The variable $L^{\rm off}(t)$ is the number of non-illuminated lanes.

Roughly speaking, Eq.~(\ref{eq:modelO}) means that when $L_{Vk}(t+1) > 0.5$, the branch in the lane $Vk$ should contract approximately by $2\Delta^{\rm out}$ if the original length $X_{Vk}(t)$ is longer than a certain value of $\sim 0.6$. On the other hand, Eq.~(\ref{eq:modelI}) means that there are three contributions to the elongation of branch in the lane $Vk$ when $L_{Vk}(t+1) \le 0.5$. The first contribution associated with $\Delta^{\rm in}$ assumes that a certain amount ($\Delta^{\rm in}$) of amoeba's body consistently leaks out from the hub part towards the lane part, and it results in the elongation of each branch by $\Delta^{\rm in}/L^{\rm off}$ via equitable distribution when the number of non-illuminated lanes is $ L^{\rm off}$. The second term associated with $O_{Vk}(t+1)$ means that a sum of the contracted portions of the illuminated branches are redistributed equally to the non-illuminated lanes. 

The elongation of branch occurs mostly by equitable distribution of the sum of the contracted amounts of  branches in illuminated lanes as well as the constant leak from the hub part to the $ L^{\rm off}$ non-illuminated lanes. However, there should be a case that all the lanes are illuminated, i.e., $L^{\rm off}=0$. In such a case, these contributions are stocked at the moment, and will be leaked to non-illuminated lanes equally when $L^{\rm off}$ becomes nonzero. The variable $S(t)$ defined in Eq.~(\ref{eq:modelS}) describes this situation, and this is the third contribution to the elongation described in Eq.~(\ref{eq:modelI}).

In addition, the following condition is imposed in the computational model to reproduce the experimental results,
\begin{equation}
\label{eq:Condition}
\sum_{Vk \in \rm{off}} \{ I_{Vk}(t+1) + \xi_{Vk}(t+1) \} 
-\sum_{Vk \in \rm{on}} \{ O_{Vk}(t+1) - \xi_{Vk}(t+1) \} \approx \Delta^{\rm in}.
\end{equation}
Here $\sum_{Vk \in {\rm off}}$ ($\sum_{Vk \in {\rm on}}$) denotes the summation over the non-illuminated (illuminated) lanes. This condition represents a law of conservation where the sum of the elongated (positive) and contracted (negative) amounts of the branches equals to $\Delta^{\rm in}$ constantly at every iteration step, which is the amount of leak from the hub part to the lanes. This condition is introduced by an idea that if the sum of $X_{Vk}$ increases by a certain constant rate per unit time, the time required for the solution search is expected to be proportional to $n$, because the sum of $X_{Vk}$ becomes $n$ in the end. The variable $S(t)$ is introduced to satisfy this condition at every iteration step as long as $L^{\rm off} \ne 0$.

In the previous simulation in Ref. \cite{Zhu18}, they perform 1000 trials of a solution search for the TSP with 10 to 20 cities by using the original Amoeba TSP algorithm. The obtained results coincide well with those in the experiment. The approximate solutions are obtained for almost all the trials and the average number of iterations turns out to be proportional to $n$. Moreover, the average of the normalized route length $R_{\rm calc}/R_{\rm est}$ takes a constant value of 0.9 irrespective of the number of cities $n$. 

It should be mentioned that although the number of iterations is proportional to $n$, the total CPU time for the calculation is, in fact, proportional to $n^5$. The serial process of CPU requires computational time proportional to $n^4$ for treating the $n^4$ components of the cost tensor $W_{Vk, Ul}$ to calculate the illumination pattern $L_{Vk}(t)$. Therefore, only the computational model is not sufficient for the practical application. We need to physically implement the model as a combinatorial optimization machine where the serial calculation by CPU is substituted with parallel physical phenomena to realize the efficient solution search. However, the physical implementation of the Amoeba TSP algorithm is difficult, and only an attempt using analog electronic circuits has succeeded in this challenge so far \cite{Saito20a}. The difficulty stems partly from the strict condition of Eq.~(\ref{eq:Condition}). In this paper, we propose a modified computational model of the amoeba-inspired combinatorial optimization machine, which, we expect, is suitable for physical implementation.
 
\section{Calculation conditions}
We present the values of parameters in the numerical simulations and how to determine them before introducing the results. 
\subsection{Parameters}
Table \ref{tab:Parameter} shows the values of parameters used in our numerical simulations. The parameters $\lambda$ and $\mu$ give costs as penalties to the revisiting once-visited city and the simultaneous visits to multiple cities, respectively. Their values are fixed at $\lambda$=0.5 and $\mu$=0.5 throughout the numerical simulations. The parameter $\nu$ gives the cost proportional to the route length. Importantly, the costs for violating the constraints must be larger than those for increasing in the route length. On the other hand, the difference between route lengths should be evaluated as large as possible. On the basis of these two policies, the value of $\nu$ is determined by the following procedure. We consider combinations of three cities ($V_1$, $V_2$, $V_3$) out of $n$ cities and the routes from $V_1$ to $V_3$ via $V_2$ where the route length is $d(V_1, V_2) + d(V_2, V_3)$. The value of $\nu$ is determined so as to satisfy the following inequality,
\begin{equation}
\label{eq:nu}
\nu \cdot \max\{d(V_1,V_2)+d(V_2,V_3) \} \leq \min\{\lambda, \mu\}.
\end{equation}
We also consider that the value of $\nu$ should be set as close as possible to $\nu^*$ where,
\begin{equation}
\label{eq:nu2}
\nu^* \equiv \frac{\min\{\lambda, \mu\}}{\max\{d(V_1,V_2)+d(V_2,V_3) \}}.
\end{equation}
For example, we set $\nu$=0.00176 when $n$=20.

The parameter $\delta$ gives upper and lower bounds of the uniform random number $\xi_{Vk}$ in the Amoeba TSP algorithm. The parameters $\Delta^{\rm out}$ and $\Delta^{\rm in}$ are fixed at $\Delta^{\rm out}$=0.001 and $\Delta^{\rm in}$=0.001 throughout the present work. 
\begin{table}[h]
\begin{center}
\caption{Parameter values used for the numerical simulations.}
\label{tab:Parameter}
$\begin{array}{cccccc}
\hline 
\lambda & \mu & \delta & \Delta^{\rm{out}} & \Delta^{\rm in}  \\
\hline
0.5 & 0.5 & 0.003 & 0.001 & 0.001 \\
\hline
\end{array}$
\end{center}
\end{table}

Maps of TSP are produced such that all the distances between two cities are given by a set of normal random numbers with a mean of 100 and a standard deviation of 17, with which $R_{\rm est}$ approximately equals to $100n$ for the $n$-city TSP. 

\subsection{Conversion to the solution}
In our numerical simulations, we abort the iteration to obtain an approximate solution $\{ X_{Vk}^{\rm bin} \}$ when the binary variables $X_{Vk}^{\rm bin} \equiv \theta(X_{Vk}-0.99)$ satisfy the following conditions,
\begin{equation}
\label{eq:sumXbin}
\sum_{V=1}^{n} X_{Vk}^{\rm{bin}}=1, \hspace{1cm} \sum_{k=1}^{n} X_{Vk}^{\rm{bin}}=1.
\end{equation}
In the iteration process, we check whether the conditions are satisfied at every iteration step.

\section{Modifications of the model}
We focus on the following three elements A, B, and C, which characterize the Amoeba TSP algorithm, and examine how their modifications affect the optimization performance,
\begin{description}
\item[(Element A)] The fluctuation $\xi_{Vk}$ is given by a uniform random number,
\item[(Element B)] The sum of $X_{Vk}$ increases by $\Delta^{\rm in}$ at every iteration step as described in Eq.~(\ref{eq:Condition}),
\item[(Element C)] Sigmoid functions are used in Eq.~(\ref{eq:expL}) for $L_{Vk}(t+1)$ and in Eq.~(\ref{eq:modelO}) for $O_{Vk}$. 
\end{description}
We perform 1000 solution-search trials for the 20-city TSP using a computational model with a modification in one of these three elements. We compare the obtained results with those obtained by the original algorithm to judge which elements are substantially important for the optimization performance.

\subsection{Modifications of Element A}
We first examine the following two types of modifications for Element A, that is, 
\begin{description}
\item[(A-1)] Removing the fluctuations by setting $\xi_{Vk}$=0,
\item[(A-2)] Using normal random numbers for the fluctuations $\xi_{Vk}$.
\end{description}
Table \ref{tab:modA} shows the results obtained by the modified models together with those obtained by the original model using uniform random numbers ranging in $[-0.003, 0.003]$ for the fluctuations. Noticeably, we cannot obtain an approximate solution in any of 1000 trials with the modification (A-1), which clearly indicates that the solution-search ability of amoeba indeed relies on the fluctuations in its dynamics. For the modification (A-2), we examine the normal random numbers with a mean of 0 and a standard deviation of 0.003 for $\xi_{Vk}$ and find that the number of iterations is significantly suppressed by 29\% as compared with the original algorithm, while the success rate does not change so much. Summarizing the results of these examinations, we conclude that the fluctuation $\xi_{Vk}$ is indispensable for the solution-search procedure, and normal random numbers are more suitable for efficient searches, which can significantly reduce the required number of iterations as compared with uniform random numbers.
\begin{table}[h]
\caption{Optimization results obtained by the modifications of Element A. The results are evaluated by focusing on the three criteria, i.e., the rate of the trials in which the modified algorithm successfully finds an approximate solution within 3000 iterations (Success rate), the average number of iterations required to find an approximate solution (Average number of iterations), and the average of the normalized route lengths of obtained solutions (Average of $R_{\rm calc}/R_{\rm est}$). The results obtained by the original Amoeba TSP algorithm is presented for comparison.} 
\label{tab:modA}
\begin{center}
\begin{tabular*}{15cm}{@{\extracolsep{\fill}}lccc}
\hline
Modification & Success rate & Average number of iterations & Average of $R_{\rm calc}/R_{\rm est}$\\
\hline
(A-1)        & 0.000 & ------    & ------ \\
(A-2)        & 0.986 & 1326.8 & 0.941 \\
\hline \hline
Original   & 0.992 & 1870.6 & 0.951 \\
\hline
\end{tabular*}
\end{center}
\end{table}

\subsection{Modifications of Element B}
We next examine the following four types of modifications for Element B, that is,
\begin{description}
\item[(B-1)] Replacing $I_{Vk}(t)$ with $0.9 I_{Vk}(t)$ in Eq.~(\ref{eq:modelX}),
\item[(B-2)] Replacing $I_{Vk}(t)$ with $1.1 I_{Vk}(t)$ in Eq.~(\ref{eq:modelX}),
\item[(B-3)] Setting $\Delta^{\rm in}$ at zero in Eqs.~(\ref{eq:modelI}) and (\ref{eq:modelS}),
\item[(B-4)] Replacing $L^{\rm off}$ with $n$ in Eq.~(\ref{eq:modelI}).
\end{description}
Table \ref{tab:modB} shows the results obtained by the modified models together with those obtained by the original model. Concerning the modifications (B-1) and (B-2), the increase rate $I_{Vk}$ of $X_{Vk}$ when $L_{Vk} \leq 0.5$ is suppressed by 10\% for the former, while it is enhanced by 10\% for the latter. The conservation law in Eq.~(\ref{eq:Condition}) is violated by these modifications. For the modification (B-3), the constant leak from the hub part to the lanes $\Delta^{\rm in}$ is set to be zero. With this modification, a sum of the elongated and contracted amounts of the branches is always cancelled except for the contributions from the fluctuations. For the modification (B-4), $L^{\rm off}$ in Eq.~(\ref{eq:modelI}) is repalced with the number of cities $n$, which is based on an idea that $L^{\rm off}$ finally becomes $n$ when a solution is obtained. This modification releases us from counting the number $L^{\rm off}$ of non-illuminated lanes at every iteration step. Moreover, this modification is expected to boost the solution search by enhancing the increase rate $I_{Vk}$ at each step because $n$ should be much smaller than $L^{\rm off}$ particularly in the early stage of the iteration.

According to Table~\ref{tab:modB}, we find that a larger increase rate $I_{Vk}$ of $X_{Vk}$ at every step results in the better performance in terms of the number of iterations and the route length. In particular, the modification (B-4) reduces the number of iterations by 44\% as compared to the original Amoeba TSP algorithm. On the contrary, the success rate is not improved by any of these modifications.

So far, it has been naively believed that nonlocal correlations among branches manifested by the conservation law are indispensable to reproduce the solution-search ability of amoeba. Namely, we assume that a sum of the elongated (positive) and contracted (negative) amounts of branches should constantly equal to the amount of leak $\Delta^{\rm{in}}$ from the amoeba body in the hub part. However, our study has demonstrated that the conservation rule is not required to solve TSP for the amoeba TSP algorithm. Moreover, it has turned out that much better solutions can be obtained if the conservation rule is removed. Surprisingly, the excellent solution-search ability of amoeba turns out to be attributable only to its fluctuating dynamics associated with its shape deformation.
\begin{table}[h]
\caption{Optimization results obtained by the modifications of Element B.}
\label{tab:modB}
\begin{center}
\begin{tabular*}{15cm}{@{\extracolsep{\fill}}lccc}
\hline
Modification & Success rate & Average number of iterations & Average of $R_{\rm calc}/R_{\rm est}$\\
\hline
(B-1)	    & 0.990 & 1937.4 & 0.957 \\
(B-2)        & 0.992 & 1817.4 & 0.949 \\
(B-3)        & 0.996 & 1989.7 & 0.958 \\
(B-4)        & 0.994 & 1049.3 & 0.912 \\
\hline \hline
Original   & 0.992 & 1870.6 & 0.951 \\
\hline
\end{tabular*}
\end{center}
\end{table}

\subsection{Modifications of Element C}
We finally examine the following three types of modifications for Element C, that is, 
\begin{description}
\item[(C-1)] Replacing the sigmoid function $\sigma_{20, 0.6}(x)$ in Eq.~(\ref{eq:modelO}) with 1,
\item[(C-2)] Replacing the sigmoid function $\sigma_{1000, -0.5}(x)$ in Eq.~(\ref{eq:expL}) with the step fumction $\theta(x+0.5)$,
\item[(C-3)] Replacing the sigmoid function $\sigma_{35, 0.6}(x)$ in Eq.~(\ref{eq:expL}) with the step fumction $\theta(x-0.6)$.
\end{description}
Table \ref{tab:modC} shows the results obtained by the modified models together with those obtained by the original model. For the modification (C-1), the decrease rate $O_{Vk}$ of $X_{Vk}$ when $L_{Vk} > 0.5$ becomes independent of $X_{Vk}$. We find that the number of iterations is reduced by 48\% as compared with the original Amoeba TSP algorithm by this modification. For the modification (C-2), the sigmoid function $\sigma_{1000, -0.5}(x)$ in Eq.~(\ref{eq:expL}) is replaced with the step function $\theta(x+0.5)$. According to Table \ref{tab:modC}, we learn that the results are not affected so much by these modifications. For the modification (C-3), the sigmoid function $\sigma_{35, 0.6}$ in Eq.~(\ref{eq:expL}) is replaced with the step function with $\theta(x-0.6)$. Table \ref{tab:modC} shows that the success rate becomes much worse with this modification, i.e., typically less than half of those obtained by the original algorithm. Moreover, the number of iterations becomes 1.4 times larger than those for the original algorithm.

These results offer several insights into the Amoeba TSP model to achieve a better optimization with less computational time. First, the decrease rate $O_{Vk}$ of $X_{Vk}$ in Eq.~(\ref{eq:modelO}) should be given as a constant $2 \Delta^{\rm out}$ without depending on the value of  $X_{Vk}$. Second, the sigmoid function $\sigma_{1000, -0.5}$ in Eq.~(\ref{eq:expL}) should be replaced with the step function. On the other hand, $\sigma_{35, 0.6}$ is necessary to search approximate solutions of the TSP. Note that the argument of $\sigma_{1000, -0.5}$, $\sum_{U=1}^n \sum_{l=1}^n W_{Vk, Ul} \sigma_{35, 0.6}(X_{Ul}(t))$, corresponds to the derivative of the cost function in Eq.~(\ref{eq:CostFunc}) by $x_{Vk}$ when $\sigma_{35, 0.6}$ is replaced with the step function with a threshold 0.6 and $x_{Vk}$ is defined as $\theta(X_{Vk}-0.6)$. Owing to the finite slope of $\sigma_{35, 0.6}$, the combination of $X_{Vk}$ resulting in larger cost, $\sum_{U=1}^n \sum_{l=1}^n W_{Vk, Ul} \sigma_{35, 0.6}(X_{Ul}(t)) (\leq -0.5)$, is sometimes allowed in the solution-searching process and thus the computational model is able to search in wider solution space without trapped in certain local optimal solutions.
\begin{table}[h]
\caption{Optimization results obtained by the modifications of Element C.}
\label{tab:modC}
\begin{center}
\begin{tabular*}{15cm}{@{\extracolsep{\fill}}lccc}
\hline
Modification & Success rate & Average number of iterations & Average of $R_{\rm calc}/R_{\rm est}$\\
\hline
(C-1)        & 1.000 & 974.5 & 0.952 \\
(C-2)        & 0.991 & 1874.8 & 0.953 \\
(C-3)        & 0.460 & 2578.7 & 1.000 \\
\hline \hline
Original   & 0.992 & 1870.6 & 0.951 \\
\hline
\end{tabular*}
\end{center}
\end{table}

\section{Improved Amoeba TSP model}
According to the above examinations, we have found three modifications that can improve the optimization results significantly,
\begin{itemize}
\item Using normal random numbers for the fluctuations $\xi_{Vk}$ (A-2),
\item Replacing $L^{\rm off}$ with the number of cities $n$ (B-4),
\item Replacing the sigmoid function $\sigma_{20,0.6}(x)$ in Eq.~(\ref{eq:modelO}) with unity (C-1).
\end{itemize}
In Table~\ref{tab:gcond}, we also summarize items of the optimization improved by each modification. We construct a computational model named Improved Amoeba TSP algorithm by applying the three modifications to the Amoeba TSP algorithm. We perform 1000 solution-search trials for the $n$-city TSPs with $n$=10-100 by using the Improved Amoeba TSP algorithm. 
\begin{table}[h]
\caption{Effects of each modification on the optimization. The symbols $++$, $+$ and $-$ denote excellent, good and bad effects for each criterion, respectively. The symbol 0 denotes a neutral effect.}
\label{tab:gcond}
\begin{center}
\begin{tabular*}{15cm}{@{\extracolsep{\fill}}lccc}
\hline
Modification & Success rate & Average number of iterations & Average of $R_{\rm calc}/R_{\rm est}$\\
\hline
(A-2)  & $-$ &$+$ & $+$\\
(B-4)  & 0 & $++$ & $+$ \\
(C-1)  & $+$ & $++$ & 0 \\
\hline
\end{tabular*}
\end{center}
\end{table}

\begin{figure*}[thb]
\includegraphics[scale=1.0]{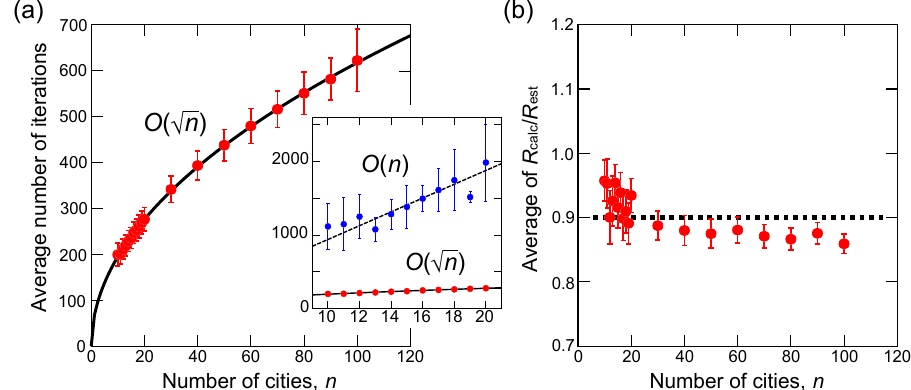}
\caption{(a) City-number dependence of the average number of iterations required to obtain an approximate solution using the Improved Amoeba TSP algorithm, which turns out to scale with $\sqrt{n}$. Inset shows the comparison with results of the original Amoeba TSP algorithm \cite{Zhu18}, which scales linearly with $n$. (b) City-number dependence of the average of normalized route length $R_{\rm calc}/R_{\rm est}$ obtained by the Improved Amoeba TSP algorithm. The dashed lines represents $R_{\rm calc}/R_{\rm est}=0.9$ as a universal result of the original Amoeba TSP algorithm.}
\label{fig:IATSP}
\end{figure*}

The results are summarized in Table~\ref{tab:IATSP}. We first learn that approximate solutions are obtained in all the 1000 trials for all the choices of $n$. When $n$=20 for example, the Improved Amoeba TSP algorithm requires reduced number of iterations to obtain an approximate solution as compared to the original Amoeba TSP algorithm. This high efficiency enables us to solve the TSP with $n$ greater than 20 within reasonable computational time. Figures \ref{fig:IATSP}(a) and \ref{fig:IATSP}(b) show the $n$ dependence of the average number of iterations and the average of $R_{\rm calc}/R_{\rm est}$, respectively. The number of iterations grows proportionally to $\sqrt{n}$. This scaling relation is nontrivial and surprising, and it holds irrespective of the distribution pattern of distances between cities. Moreover, the average of $R_{\rm calc}/R_{\rm est}$ decreases gradually but monotonically as $n$ increases, and is expected to be comparable to or even smaller than that for the original algorithm when $n$ is large enough. These results indicate that the Improved algorithm is superior to the original algorithm in terms of the efficiency and certainty of the solution search.
\begin{table}[h]
\caption{Optimization results obtained by the Improved Amoeba TSP algorithm.}
\label{tab:IATSP}
\begin{center}
\begin{tabular*}{15cm}{@{\extracolsep{\fill}}cccc}
\hline
Number of cities $n$ & Success rate & Average number of iterations & Average of $R_{\rm calc}/R_{\rm est}$ \\
\hline
10 & 1.00     & 199.5 & 0.957 \\
11 & 1.00     & 201.3 & 0.953 \\
12 & 1.00     & 211.1 & 0.900 \\
13 & 1.00     & 219.1 & 0.926 \\
14 & 1.00     & 229.0 & 0.954 \\
15 & 1.00     & 235.8 & 0.916 \\
16 & 1.00     & 247.0 & 0.939 \\
17 & 1.00     & 253.2 & 0.899 \\
18 & 1.00     & 260.4 & 0.910 \\
19 & 1.00     & 269.3 & 0.891 \\
20 & 1.00     & 276.3 & 0.934 \\
30 & 1.00     & 341.5 & 0.887 \\
40 & 1.00     & 393.3 & 0.880 \\
50 & 1.00     & 437.7 & 0.875 \\
60 & 1.00     & 479.5 & 0.881 \\
70 & 1.00     & 515.9 & 0.871 \\
80 & 1.00     & 550.6 & 0.867 \\
90 & 1.00     & 581.4 & 0.876 \\
100 & 1.00   & 622.2 & 0.859 \\
\hline
\end{tabular*}
\end{center}
\end{table}

\section{Conclusion}
To conclude, we proposed an improved computational model of the amoeba-inspired combinatorial optimization machine, named Improved Amoeba TSP model, through examining the original model proposed by Aono $et$ $al$. First, we focused on three elements in the original model and examined how each of these elements affects the optimization performance. Appropriate modifications of these elements were found to significantly improve the optimization results. Importantly, it turned out that the volume conservation law assumed in the original model is not necessarily required to realize the high solution-search ability in contrast to our naive belief. This justifies the modification, which might release us from possible related constraints and restrictions against the physical implementation. Next, we constructed the Improved Amoeba TSP model by applying these three modifications to the original model. The proposed model indeed provides noticeably improved results in terms of the number of iterations and the route length. In particular, we found that the number of iterations scales to order $\sqrt{n}$ for the number of cities $n$, which is significantly suppressed as compared to the scaling to order $n$ for the original model. Our study offers the guides to enhancing the solution-search ability of the amoeba-inspired combinatorial optimization machine and gives a basis towards their implementations with physical devices.

\section{Acknowledgment}
We are indebted to Masashi Aono for insightful discussions. This work is supported by Japan Society for the Promotion of Science KAKENHI (Grant No.~20H00337, 22H05114), CREST, the Japan Science and Technology Agency (Grant No.~JPMJCR20T1), and a Waseda University Grant for Special Research Projects (Project No.~2023C-140, 2023E-026, 2024C-153, 2024C-155).

\end{document}